\documentclass{article}

\usepackage{PRIMEarxiv}
\usepackage[utf8]{inputenc} %
\usepackage[T1]{fontenc}    %
\usepackage{hyperref}       %
\usepackage{url}            %
\usepackage{booktabs}       %
\usepackage{amsfonts}       %
\usepackage{nicefrac}       %
\usepackage{microtype}      %
\usepackage{lipsum}
\usepackage{fancyhdr}       %
\usepackage{graphicx}       %
\graphicspath{{media/}}     %

\usepackage{bbm}
\usepackage{amssymb}
\usepackage[usenames, dvipsnames]{color}
\usepackage{amsmath}
\usepackage[numbers]{natbib}
  
\title{SCALES: From Fairness Principles to Constrained Decision-Making
\thanks{\textit{Proceedings of the 2022 AAAI/ACM Conference on AI, Ethics, and Society}} 
}

\author{
  Sreejith Balakrishnan$^{+}$, Jianxin Bi$^{+}$, Harold Soh\\
  National University of Singapore \\
  Singapore \\
  \texttt{\{sreejith,jianxin.bi,harold\}@comp.nus.edu.sg}\\
  $^{+}$ \textit{Both authors contributed equally to this research.}
}

\DeclareMathOperator{\argmaxH}{argmax}

\newcommand{\R}{\mathbb{R}}
\newcommand{\E}{\mathbb{E}}
\newcommand{\scales}{\textsc{Scales}}

\begin{document}
\maketitle

\begin{abstract}
This paper proposes SCALES, a general framework that translates well-established fairness principles into a common representation based on the Constraint Markov Decision Process (CMDP). With the help of causal language, our framework can place constraints on both the procedure of decision making (procedural fairness) as well as the outcomes resulting from decisions (outcome fairness). Specifically, we show that well-known fairness principles can be encoded either as a utility component, a non-causal component, or a causal component in a SCALES-CMDP. We illustrate SCALES using a set of case studies involving a simulated healthcare scenario and the real-world COMPAS dataset. Experiments demonstrate that our framework produces fair policies that embody alternative fairness principles in single-step and sequential decision-making scenarios.   
\end{abstract}

\section{Introduction}

Algorithmic decision-making has become increasingly popular in real-world applications across numerous domains including  healthcare~\cite{kleinberg2015prediction, patel2002emerging, abdelaziz2017intelligent}, banking~\cite{huang2007credit, kumar2020review}, human resources~\cite{chalfin2016productivity}, and public policy~\cite{kleinberg2016inherent,brayne2017big,lowenkamp2009development,hurley2018can,toros2018prioritizing}. Unfortunately, many proposed algorithms focus on narrow performance criteria and disregard broader societal concerns such as \emph{fairness}; these methods do not ensure fairness in either their \emph{procedure} or in the resulting \emph{outcomes}. There is a risk that when deployed at scale, algorithms oblivious to fairness considerations will magnify injustices that tear at the fabric of society. 
Recognition of this issue has given rise to an active research area that defines and incorporates fairness into decision-making~\cite{verma2018fairness,castelnovo2021zoo}.

\begin{figure}
  \centering
  \includegraphics[width=0.6\linewidth]{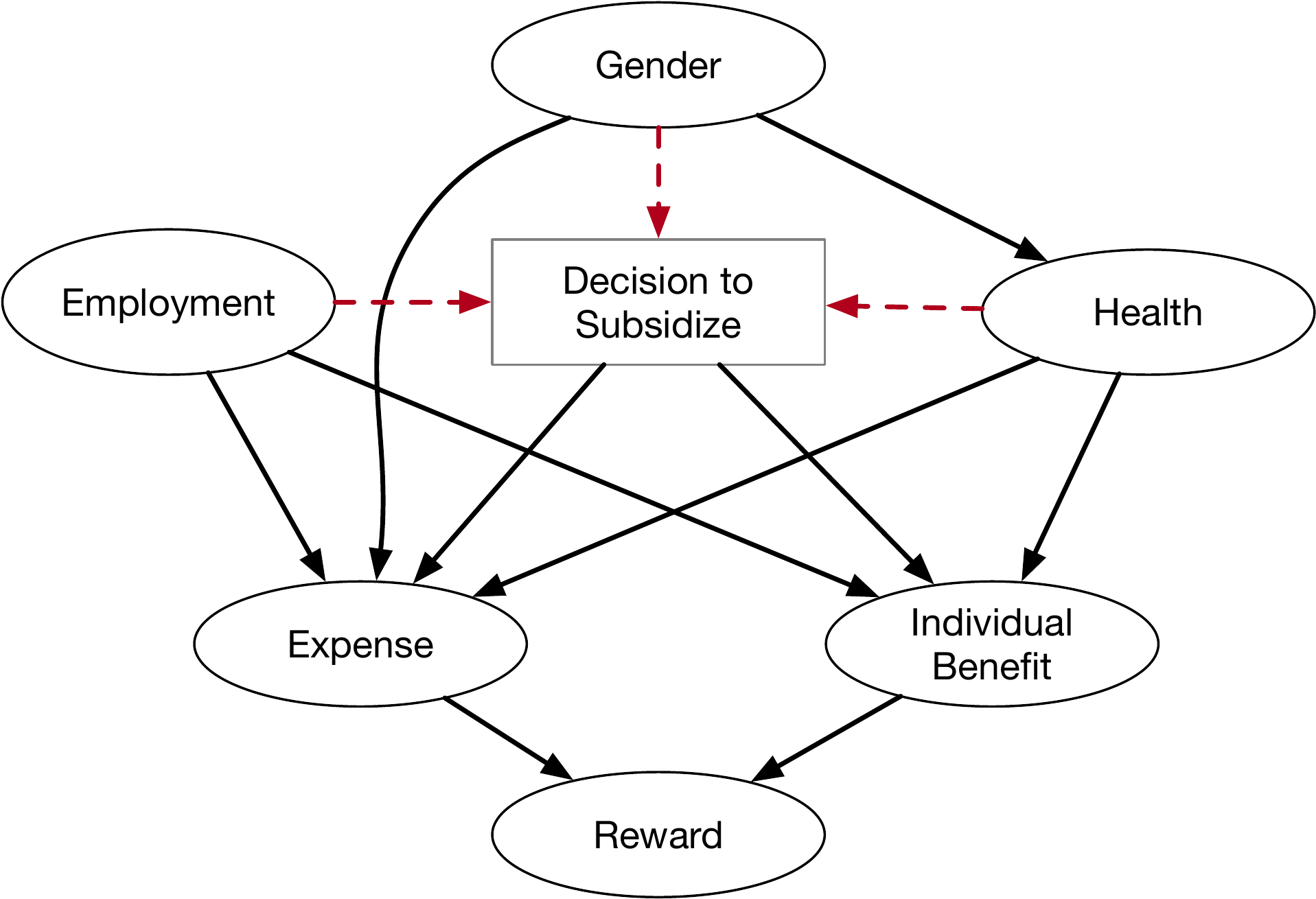}
  \caption{Causal Model used by a government to decide on health care subsidies. Each citizen is defined using their gender, current health, and employment status. The model also shows the effect of the decision on the individual and the government. The gender of the individual does not \textit{directly} affect his/her reward, but it does have a direct causal relationship with the cost incurred by the government.}
  \label{fig:causalmodel}
\end{figure}

One notable finding from this on-going research is that there is no single definition of fairness --- the fairness principles that are adopted reflect the social values of decision-makers. 
Consider the scenario illustrated in Figure \ref{fig:causalmodel}. Here, we show the causal model employed by a fictitious Government to render a decision on whether to provide subsidized health care to a citizen (identified by his/her gender, health condition, and employment status). The government tries to achieve a balance between the welfare of citizens and the cost incurred by the subsidy, which we will refer to as a total ``reward''. In such a setting, different fairness criteria can apply:
\begin{itemize}
    \item \textbf{Counterfactual fairness in decisions} ~\cite{kusner2018counterfactual}. A fair government might be interested in ensuring that its decisions are not influenced by sensitive attributes such as gender. 
    \item \textbf{Group Fairness}~\cite{chouldechova2016fair,kamiran2012,hardt2016equality}. The government tries to ensure that the benefits enjoyed by the citizens are well balanced across people in different gender groups.
    \item \textbf{Path-specific fairness} \cite{wu2019pcfairness,chiappa2019path}. The government might instead want to make sure that gender only affects their decisions through non-discriminatory factors such as a citizen's health. 
\end{itemize}
The above are all valid fairness principles and prior work has largely focused on methods for defining and computing \emph{specific} fairness principles for fair decision-making. However, 
humans are notoriously poor at designing objectives for machines and policymakers may not agree on which fairness principles apply in a given scenario, nor appreciate what the resultant policies would be.

To bridge this gap, we need a flexible decision-making framework that can incorporate alternative fairness principles in a standard manner. Such a tool enables the principled derivation of policies and can help decision-makers better understand what policies emerge from a set of fairness principles. In our health subsidy example, the three different fairness criteria above could be critically examined with respect to individual health benefits and total expenses incurred.

This paper presents the \textbf{S}tructured \textbf{CA}usal \textbf{L}anguage \textbf{E}ffects \textbf{S}ynthesis (\scales) framework for fair decision-making. 
\scales{} casts fairness principles as constraints in a Constrained Markov Decision Process (CMDP)~\cite{altman1999constrained}. We focus our discussion on how one can represent fairness principles as constraints and show that fairness principles can be translated into a combination of causal, non-causal, and utility components. While the non-causal and utility constraints are based on the outcome of a decision, specifying the causal constraints requires the language of causal inference. Specifically, the causal effects implied by various fairness principles are explicitly represented in our framework using Path-specific Counterfactual Effects (PCE), a general representation of causal effects~\cite{wu2019pcfairness}. Using our above example of subsidized healthcare, we will show in Sec. \ref{sec:casestudiestitle} that all three fairness principles above (and others) can be stated in a clean, coherent manner within \scales. 

One key advantage of our framework lies in its transparency relative to existing works~\cite{nabi2018fair,nabi2019learning, zhang2016causal, chiappa2020general, kilbertus2017avoiding, zhang2018fairness}. 
For example, \citet{nabi2019learning} and \citet{zhang2016causal} have proposed indirect multi-step optimization processes that modify the causal mechanism of the environment to generate a counterfactual world where all decisions are fair. %
In contrast, \scales{} directly maximizes a constrained policy using a world model. Direct optimization circumvents the additional task of transferring samples from the observed distribution to their fair-world counterparts, which remains an open problem. \scales{} does require the use of capable solvers and we find that existing solvers such as CPO~\cite{achiam2017constrained} work sufficiently well in our experiments. In particular, our results indicate that \scales{} enables the direct comparison of various fairness criteria and their corresponding fair policies under a common modeling framework.

\subsection{Summary.} This paper makes the following three key contributions:
\begin{itemize}
    \item \scales, a coherent framework for fair decision-making based on CMDPs;
    \item A comprehensive list of fairness principles and their corresponding PCE representations;
    \item Case studies that illustrate how \scales{} can be used to learn fair optimal policies under different fairness principles.
\end{itemize}
More broadly, we hope that this paper serves as a bridge between the fairness and (constrained) decision-making communities.
\scales{} favors paradigm clarity, which we believe is crucial for trustworthy deployment~\cite{kok20trust,xie2019robot}. We believe our formulation also naturally leads to avenues for future research, such as better solvers that exploit structure in the constraints. \scales{} also forms a foundation for extensions that consider partial-observability and multiple conflicting objectives~\cite{sohmrpomdp}, which are characteristic of real-world decision-making.

\section{Preliminaries}
At a high level, \scales{} is a formalization of fair decision-making using specific objectives and constraints (including those that consider causal effects). The following is a brief introduction to constrained sequential decision-making, causal modeling~\cite{pearl2009causality} and fairness principles. We focus on the key essentials on fairness and refer readers desiring more detail to comprehensive surveys~\cite{10.1145/3194770.3194776,castelnovo2021zoo}.

\subsection{Constrained Markov Decision Process}
\subsubsection{Markov Decision Process (MDP)} An MDP models a sequential decision making scenario where an agent is taking actions in a world to maximize rewards. More formally, it is defined by a tuple $\mathcal{M} : \langle S, A, \mathcal{T}, R, \gamma \rangle$ where $S$ is a finite set of states, $A$ is a finite set of actions, $\mathcal{T}(s,a,s') = P(s'|s,a)$ is the conditional probability of next state $s'$ given current state $s$ and action $a$, $R:S\times A\times S \rightarrow \displaystyle \R$ denotes the reward function, and $\gamma \in (0,1)$ is the discount factor. An optimal policy $\pi^*$ is a policy that maximizes the expected sum of discounted rewards:
\begin{align} 
J_{R}(\pi) = \E_{\tau \sim \pi}\left[\sum_{t=0}^{H-1} \mathcal{\gamma}^t R(s_t, a_t, s_{t+1}) |\pi,\mathcal{M}\right].
\end{align}
Here $\tau$ is a trajectory of horizon $H$ consisting of state action sequence $(s_0,a_0,s_1..,s_H)$ sampled from the distribution of trajectories induced by policy $\pi$. The task of finding an optimal policy is referred to as policy optimization. 

\subsubsection{Constrained MDP (CMDP)} A CMDP~\cite{altman1999constrained} is an extended version of MDP that imposes  constraints on the allowable set of policies. In addition to the standard components of an MDP, a CMDP specifies $\mathcal{C} \triangleq \{C_i \}_{i=1}^M$, a set of $M$ cost functions $C_i: S \times A \times S \rightarrow \displaystyle \R$. 
The expected discounted cost value for a cost function $C_i$ is given by
\begin{align} 
J_{C}^{i}(\pi) = \E_{\tau \sim \pi}  \left[\sum_{t=0}^{H-1} \mathcal{\gamma}^t C_{i}(s_t, a_t,s_{t+1}) |\pi,\mathcal{M}\right]
\end{align}
Each of the cost function $C_i$ has a corresponding limits $d_i$ for $i = [1...M]$ which defines the optimal policy $\pi*$ as:

\begin{equation}\label{eq:cmdp-obj}
\begin{aligned}
\pi^{*} = \displaystyle \argmaxH_{\pi}& \hspace{-2.4mm}& &J_R(\pi)\\
\hspace{-2.4mm} \displaystyle \text{s.t. }& \hspace{-2.4mm}& &J_{C}^{i}(\pi) < d_i, i = 1, \dots, M
\end{aligned}
\end{equation}

CMDPs are used extensively in Safe Reinforcement Learning (RL) \cite{garcia2015comprehensive,achiam2017constrained} where safety considerations are formalized as constraints to be satisfied while learning a policy and/or during deployment. We will see that Fair Reinforcement Learning can also be cast as a specialized CMDP. However, fairness considerations are different from safety, e.g., when we consider counterfactuals or the fair distribution of resources. In later sections, we will show that fairness constraints will have specific structure depending on the principle applied. %

\subsection{Causal Modeling}
\subsubsection{Structural Causal Models and Causal Graphs}
A causal graph  $\mathcal{G} = \langle \mathcal{V}, \mathcal{E} \rangle$ is a Directed Acyclic Graph (DAG) with nodes $\mathcal{V}$ and edges $\mathcal{E}$~\cite{pearl2009causality}. An edge $e \in \mathcal{E}$ from node $v_1 \in \mathcal{V}$ to $v_2 \in \mathcal{V}$ indicates a direct causal association between $v_1$ and $v_2$ with $v_1$ determining the value of $v2$. For instance, in Figure \ref{fig:causalmodel} the gender of a person has a direct causal effect on their health. While a causal graph allows one to understand the flow of causal and confounding associations~\cite{pearl2009causality}, it does not explicitly describe the generating process. For that, we need a structural causal model.

A structural causal model (SCM) consistent with a causal graph $\mathcal{G}$ is specified by $\mathcal{M_G}: \langle \mathbf{U}, \mathbf{V}, \mathbf{F}, P(\mathbf{U}) \rangle$ where
\begin{itemize}
    \item $\mathbf{U}$ is a set of exogenous variables not modelled in $\mathcal{M_G}$
    \item $P(\mathbf{U})$ is the joint probability distribution defined over $\mathbf{U}$
    \item $\mathbf{V}$ is a set of endogenous variables whose values are determined by the causal mechanism specified by $\mathcal{M_G} $
    \item $\mathbf{F}$ is a set of functions mapping $\mathbf{U} \times \mathbf{V} \rightarrow \mathbf{V}$. In particular, the causal mechanism of a variable $V \in \mathbf{V}$ is specified by $v = f_{V}(\mathbf{pa}_V,u_V)$ where $f_V \in \mathbf{F}$; $u_V$ is an instance of $U_V \in \mathbf{U}$; $\mathbf{pa}_V$ is an instance of $\mathbf{Pa}_V \in \mathbf{V}$ \textbackslash $V$. $U_V$ and $\mathbf{Pa}_V$ are sets of nodes that directly determines the value of $V$. $\mathbf{Pa}_V$ is referred to as the parents of $V$.
\end{itemize}

The SCM associated with the causal graph in Figure \ref{fig:causalmodel} has $\mathbf{V}$ = \{\textit{Gender}, \textit{Health}, \textit{Employment Status}, \textit{Decision}, \textit{Expense}, \textit{Individual Benefit}, \textit{Reward}\}; $\mathbf{U}$ comprises nodes not shown in the graph which contribute to the stochastic nature of the model. Representing the causal mechanism in the form of a causal graph or SCM enables reasoning about causal effects between nodes.

\subsubsection{Causal Effects and Path Specific Causal Effects}
Causal effects quantitatively measure the change in value of an endogenous variable $Y \in \mathbf{V}$ in a given SCM $\mathcal{M_G}$ or causal graph $\mathcal{G}$ when an intervention is performed on a treatment variable $X \in \mathbf{V}$.\footnote{In general, X can be a set of treatment variables, but here, we consider a single treatment variable for expositional simplicity.} An intervention on $X$ forces a change to the value of $X$ from $x_0$ to $x_1$ regardless of its causal mechanism $f_x$. Such an intervention is mathematically represented by the \textit{do}-operator. %

Consider an SCM $\mathcal{M_G}$ with its standard component. Let $X \in \mathbf{V}$ be the treatment variable whose value is changed from $x_0$ to $x_1$ via intervention, and $Y \in \mathbf{V}$ be the outcome variable of interest. Let us assume we observe $\mathbf{O} \in \mathbf{V}$ \textbackslash ${X,Y}$ to have some observations $\mathbf{o}$ in the factual world. Note that $X$ may have a causal relation to $Y$ through multiple paths (directly or through resolving variables). We are interested in calculating the causal effect of $X$ on $Y$ only through a specific path $\sigma$. %
The Path Specific Counterfactual Effect(PCE)~\cite{wu2019pcfairness} is defined as:
\begin{equation} \label{eq:pce}
    PCE_{\sigma}(y | x_0 \rightarrow x_1, \mathbf{o}) = \E[y_{(x_1|\sigma),(x_0|\tilde{\sigma})} | \mathbf{o}] - \E[y_{x_0} | \mathbf{o}]
\end{equation}
 
Here, $y_{(x_1|\sigma),(x_0|\tilde{\sigma})}$ is the post-intervention value of $Y$ where the values of $X$ and its descendants along path $\sigma$ are calculated using the intervention value $x_1$ while descendants along other paths $\tilde{\sigma}$ are calculated using the intervention value $x_0$. $y_{x_0}$ is the post-interventional value of $Y$ calculated using $X=x_0$ through all causal paths from $X$ to $Y$. If $\mathbf{O} = \emptyset$, then we calculate the path-specific causal effect ~\cite{wu2019pcfairness,chiappa2019path} of $X$ on $Y$. Finally if $\sigma$ contains all the causal paths from X to Y, then we calculate the Total Causal Effect~\cite{wu2019pcfairness} of $X$ on $Y$. 

\subsection{Fairness Principles}
Numerous fairness principles have been studied across various research communities, from machine learning to economics~\cite{DBLP:journals/corr/abs-2010-09553,mitchell2021algorithmic,castelnovo2021zoo,10.1145/3457607,verma2018fairness,loftus2018causal}. Here, we categorize existing fairness principles into four main groups: 
\begin{itemize}
    \item \textbf{Distributive Fairness} is concerned with fair distribution of limited resources across multiple stakeholders~\cite{censor1977pareto}. For example, Rawl's Maximin principle~\cite{rawls1999theory} prescribes that one should maximize the minimum payoff, i.e., emphasis is placed on the least well-off among a set of individuals. In contrast, Equity Theory~\cite{denhardt2018managing} assigns rewards to individuals in proportion to their contributions. In welfare economics, the Pigou-Dalton principle~\cite{moulin2004fair} prefers equitable allocations (based on transfers) and places conditions on social welfare functions that rank the desirability of social states. 
    \item \textbf{Group Fairness} (GF) refers to fairness principles that dictate an equal probability of an outcome\footnote{For the remainder of this section, we will use the term \textit{outcome} to refer to both decisions as well as the outcome that results from these decisions.} across different groups in the population separated by a sensitive attribute. Examples include Demographic Parity~\cite{chouldechova2016fair} and Conditional Demographic Parity~\cite{kamiran2012} that require an equal probability of being assigned a positive outcome. In cases where the ground-truth outcome is known (such as in supervised learning), Equalized Odds ~\cite{hardt2016equality} and Equality in Opportunity ~\cite{hardt2016equality,chouldechova2016fair} add additional constraints on false positive and true negative distributions of the predicted outcome. 
    \item \textbf{Individual fairness} (IF) ~\cite{dwork2011fairness} encodes the notion that comparable individuals should be treated equally. This is usually achieved via a \emph{similarity measure} between individuals. For example, \citet{verma2018fairness} proposes Fairness through Unawareness (FTU) which requires that sensitive attributes are not used when determining individual similarity. Luck Egalitarianism (a.k.a. Equal Opportunity For Welfare)~\cite{knight2013luck, Arneson1997-ARNEAE-7} requires that no individuals are worse-off relative to others due to involuntary choices or attributes beyond one's control. Thus, it prescribes that the similarity measure should exclude features related to these attributes. 
    \item \textbf{Causality-based fairness} aims to ensure causal effect of sensitive attributes on outcome is at tolerable level. This notion of fairness has gained popularity in recent years due to (i) a growing concensus that addressing fairness requires analysis of causal effects, and (ii) technical advances in causal modeling and inference. As examples, Counterfactual Fairness~\cite{kusner2018counterfactual} leverages an SCM to study the change in outcome of an individual in a counterfactual world where a sensitive feature of the individual is altered;  \citet{kilbertus2017avoiding} investigates Unresolved Discrimination that arises due to a direct causal path between a sensitive attribute and the outcome. Many causal-based fairness principles can be unified under the umbrella of Path-Specific Counterfactual Fairness (PC Fairness)~\cite{wu2019pcfairness}. \scales{} uses PC Fairness to represent the causal aspects of fairness.  %
\end{itemize}

\section{\textsc{Scales} for Fair Decision-Making}
In this section, we introduce  \textbf{S}tructured \textbf{CA}usal \textbf{L}anguage \textbf{E}ffects \textbf{S}ynthesis (\scales), a framework that enables the formal representation of a fair decision-making problem. We use the CMDP as our base model, but formally define \scales{} CMDP (SCMDP) that places emphasis on key elements required to adequately incorporate fairness principles. For example, CMDPs do not segregate the decisions and outcomes assigned to multiple stakeholders in a given task. However, these are crucial for evaluating distributive fairness principles such as Equity Theory. 

We define a SCMDP as a tuple, $$\mathcal{M} : \langle S, A, \mathcal{T}, \mathcal{R}_e, f_w, R_a, \mathcal{C},\gamma \rangle$$ where the $S, A, \mathcal{T},$ and $\gamma$ components on a SCMDP follows the standard CMDP definitions. However, unlike the standard CMDP, the SCMDP incorporates the notion of multiple stakeholders and requires the specification of two types of reward functions: 
\begin{enumerate}
    \item $\mathcal{R}_e$, a list of individual rewards for $N$ stakeholders,
    \item $R_a $, a reward shared among the stakeholders (e.g., resource depletion, reward assigned to the decision-maker, etc.).
\end{enumerate} Let us define $\mathcal{R}_e \triangleq\{R_e^i\}_{i=1}^N$ where each $R_e^i:S\times A\times S \rightarrow \displaystyle \R $ is the reward of the stakeholder $i$. We abuse notation slightly by defining $\mathcal{R}_e(s,a,s') = \{R_e^i(s,a,s')\}_{i=1}^N$. 
The total reward $R$ is calculated by: $R(s,a,s') = f_w(\mathcal{R}_e(s,a,s')) + R_a(s,a,s')$ where $f_w$ is a social welfare function that dictates how the individual reward functions $R_e^i \in \mathcal{R}_e$ are aggregated to a real number. For example, the social welfare function to encode the Maximin principle would be $f_w(\mathcal{R}_e(s,a,s')) = \min_i(R_e^i)$.

Finally, $\mathcal{C}$ is a set of costs that along with the fixed thresholds define the constraints on the policy space. 
In the following section, we will elaborate on how fairness principles can be mapped to elements within the SCMDP, particularly the cost set $\mathcal{C}$.

\subsection{Translating Fairness Principles into SCMDP Components}

We propose a methodology that represents a given fairness principle using a combination of three components, namely 1) a utility component, 2) a non-causal component and 3) a causal component. A comprehensive table of common fairness principles and their associated components are shown in Table \ref{tab:bigtable}.

\subsubsection{Utility component.} 
Certain fairness principles require simultaneous comparisons of the utilities of multiple individuals~\cite{zimmer2021learning}. These types of fairness are incorporated into the SCMDP by specifying an appropriate reward $\mathcal{R}_e$ and social welfare function $f_w$.  Using the example above, if we seek to employ Maximin fairness 
when distributing a fixed resource across $N$ individuals, 
the social welfare function computes the minimum utility at every time step. 

\subsubsection{Non-causal component.} The non-causal component of a fairness principle are constraints that can be calculated from data (trajectories) generated by the current policy $\pi$, without any knowledge of the causal mechanism behind the decision making scenario. We seek to bound such non-causal components,
\begin{equation}\label{eq:noncausal}
    C_{NC}(\pi) < d_{NC}
\end{equation} 
where $C_{NC}$ is expressed in terms of information from the trajectories collected and $d_{NC}$ is a threshold limit set by the decision maker. For example, encoding Equity theory results in a constraint that captures the difference in ratio of an individual's reward to contribution across the population, i.e., for a given measure of contribution $b_i$, $C_{NC} = \displaystyle \E[ (R_e^{i}]/b_{i}) -  (\E[R_e^{j}]/{b_{j})} \, \forall j \neq i$. Here, we assume that the contribution $b_i$ is represented in the state.

\subsubsection{Causal component.} To obtain the causal constraints relevant to a fairness principle, we require knowledge of the causal mechanisms underlying the decision-making scenario at hand. Given a fairness principle, the task is to select the correct treatment variable, outcome variable, paths, and observations to be used in the PCE (Eq. \ref{eq:pce}) evaluated at the current policy $\pi$. Sec. \ref{sec:casestudy1} gives a specific example of a causal constraint in the healthcare subsidy scenario. To provide appropriate constraints, certain fairness principles require the application of a transformation function $f_{PCE}$ to the PCE, which results in the following constraint:
 \begin{equation}\label{eq:causalc}
     f_{PCE}(PCE_{\sigma}(y|x_0 \rightarrow x_1,\mathbf{o})) < d_{C}
 \end{equation}
In this work, we assume that there are no unobserved confounders jointly affecting any two or more endogenous variables. If the SCM $\mathcal{M_G}$ is fully-specified, we can use the causal mechanisms $f_V$ 
in the post-interventional distribution. If the SCM is not fully-specified, we can apply methods such as Inverse Probability Weighting \cite{nabi2019learning} to estimate the causal effect. If unobserved confounders are present, \citet{wu2019pcfairness} provides a method to calculate the upper bound of a causal effect; this can  serve as a suitable estimate for the PCE. 

\begin{table*}[t]\small
  \caption{Parameters of \scales{} for common fairness principles. We indicate the parameters required to specify the causal component, non-causal component, and utility component corresponding to fairness principles. In the following, $\mathbf{Z}$ is the set of sensitive attributes, and $\mathbf{\bar{Z}}$ is the set of non-sensitive attributes. $\sigma^{X,Y}$ is the set of all causal paths from nodes in $X$ to the outcome variable $Y$ and includes both direct and indirect causal paths. $\sigma^{X \rightarrow Y}$ is the set of all direct causal paths from the treatment variable(s) $X$ to an outcome variable $Y$. $\mathcal{D}$ be the node corresponding to the decision generated by the policy. $\mathcal{R}$ be the node corresponding to the outcome of a decision. Each stakeholder's contribution $b_i$ is represented in the state.}
  \label{tab:bigtable}
  \begin{tabular}{llccccccc}
  \hline
\textit{Description} &\textit{Fairness Principle} &$Y$ & $X$ & \textit{f$_w(\cdot)$} &\textit{$f_{PCE}(\cdot)$}&\textit{O} & \textit{$\sigma$} &\textit{$NC$}\\
  \hline\\
    \parbox{0.18\linewidth}{Outcome for an individual does not change when the sensitive attribute is intervened upon} & \parbox{0.14\linewidth}{Individual outcome fairness} &  $\mathcal{R}$ & $\mathbf{Z}$ & $\sum_{i=1}^N (\cdot)$ &$|\cdot|$&$\mathbf{\bar{Z}}$ & $\sigma^{\mathbf{Z},\mathcal{R}}$ & $\emptyset$\\
    \\
    \parbox{0.18\linewidth}{No total causal effect of a sensitive attribute on outcome} & \parbox{0.14\linewidth}{Group outcome fairness} &  $\mathcal{R}$ & $\mathbf{Z}$ & $\sum_{i=1}^N (\cdot)$ &$|\cdot|$&$\emptyset$ & $\sigma^{\mathbf{Z},\mathcal{R}}$ & $\emptyset$\\
    \\
    \parbox{0.18\linewidth}{No total causal effect of a sensitive attribute on decision} & \parbox{0.14\linewidth}{Group procedural fairness} &  $\mathcal{D}$ & $\mathbf{Z}$ & $\sum_{i=1}^N (\cdot)$ &$|\cdot|$ & $\emptyset$& $\sigma^{\mathbf{Z},\mathcal{D}}$  & $\emptyset$\\
    \\
    \parbox{0.18\linewidth}{Individuals are not worse-off due to uncontrollable attributes} & \parbox{0.14\linewidth}{Luck Egalitarianism} &  $\mathcal{R}$ & $\mathbf{Z}$ & $\sum_{i=1}^N (\cdot)$  &$1*(\cdot)$ &$\emptyset$ &  $\sigma^{\mathbf{Z},\mathcal{R}}$  & $\emptyset$\\
    \\
     \parbox{0.18\linewidth}{Decision for each individual does not change if the sensitive attribute changes} & \parbox{0.14\linewidth}{Individual procedural fairness} &  $\mathcal{D}$ & $\mathbf{Z}$ & $\sum_{i=1}^N (\cdot)$  & $|\cdot|$&$\mathbf{\bar{Z}}$& $\sigma^{\mathbf{Z},\mathcal{D}}$ &$\emptyset$ \\
    \\
    \parbox{0.18\linewidth}{Improve the worst payoff among a group of individuals} & \parbox{0.14\linewidth}{Maximin} &  $\emptyset$ & $\emptyset$ & $\min_{i \in N} (\cdot)$  &-&$\emptyset$& $\emptyset$ & $\emptyset$\\
    \\
     \parbox{0.18\linewidth}{Decisions should not depend directly on sensitive attributes} & \parbox{0.14\linewidth}{Path-specific Procedural Fairness} &  $\mathcal{D}$ & $\mathbf{Z}$ & $\sum_{i=1}^N (\cdot)$ & $|\cdot|$ &$\emptyset$& $\sigma^{Z\rightarrow\mathcal{D}}$ & $\emptyset$\\
     \\
      \parbox{0.18\linewidth}{Rewards should be proportional to the individual's contribution} & \parbox{0.14\linewidth}{Equity theory} &  $\emptyset$ & $\emptyset$ & $\sum_{i=1}^N (\cdot)$  &-& $\emptyset$  & $\emptyset$ & $\left|\frac {E_{\pi}[R^i_{e}]}{b_i} - \frac{E_{\pi}[R^j_{e}]}{b_j}\right|, \forall i,j$\\
    \\
    \hline
  \end{tabular}
\end{table*}

\subsection{Optimizing for Fair Policies}
Finally, we bring together elements above into a single problem formulation for fair decision-making. Given a SCMDP with specified causal, non-causal, and utility components, we seek an optimal constrained policy $\pi^*$:

\begin{equation}
    \begin{aligned}
        \pi^* = &\ \displaystyle \argmaxH_{\pi} & \hspace{-2.4mm} &J_{R|R_e,R_a,f_w}(\pi)\\
        &\ \displaystyle \textrm{such that} & \hspace{-2.4mm}&f_{PCE}(PCE_{\sigma}(y|x_0 \rightarrow x_1,\mathbf{o})) < d_{C},\\
       & & \hspace{-2.4mm}&C_{NC}(\pi) < d_{NC}
    \end{aligned}
\end{equation}

\section{Relationship to other Frameworks}

Our work is related to a growing literature on generic frameworks for fair decision-making. The salient differences are in the setting (sequential decision-making) and formulation (CMDP). Prior work~\cite{nabi2018fair,zhang2016causal,chiappa2020general} have proposed frameworks for classification and regression. 

The key idea in \cite{nabi2018fair,zhang2016causal} is to first perform \textit{discrimination discovery} --- investigating the causal structure underlying given data to identify unfair causal mechanisms --- and then replacing the unfair mechanisms to generate a \textit{fair distribution} that satisfies a given fairness principle. More recently, \citet{nabi2019learning} extended this approach to learning optimal fair policies in sequential decision-making. \citet{nabi2018fair} and \citet{nabi2019learning} use Inverse Probability Weighting (IPW) to identify the unfair mechanisms before modifying them to remove causal effects. However, this technique may not preserve fairness under the actual world distribution. Consider a scenario where we seek to minimize the causal effect of \textit{Gender} on \textit{Decisions} in the healthcare subsidy example (Figure \ref{fig:causalmodel}). We use the following notation for the various components of this example: $G$: Gender, $E$: Employment, $H$: Health, $D$: Decision, $R$: Total Reward. The causal effect of interest is given by:
\begin{flalign} \label{eq:pse_expectation}
CE_{p(Z)} &\triangleq \mathbb{E}_{p(Z)}\left[ D | do(G=1)\right] -  \mathbb{E}_{p(Z)}\left[ D | do(G=0)\right]\\
&= \sum_{D,H,E} p(D|G=1,H,E)p(H|G=1)p(E) D \nonumber\\ &\,\,- \sum_{D,H,E} p(D|G=0,H,E)p(H|G=0)p(E) D
\end{flalign}
where $Z = (G,E,H,D,R)$ and $p(Z)$ is the joint probability distribution. The interesting observation from Eq. \ref{eq:pse_expectation} is that the causal effect is independent of the probability of P(G). Hence, any new distribution generated by \textbf{only} changing P(G) will still be unfair if the original distribution is unfair. 
In addition, the IPW estimate of Eq. \ref{eq:pse_expectation} is given by
\begin{flalign} \label{eq:CE_sum_ipw}
CE_{p(Z)} \approx \frac{1}{N}\sum_{i=1}^N\left[\frac{\mathbbm{1}(G_i=1)D_i}{p(G=1)} - \frac{\mathbbm{1}(G_i=0)D_i}{1-p(G=1)}\right]
\end{flalign}
where $D_i$ are $N$ samples generated from $p(Z)$. From Eq. \ref{eq:CE_sum_ipw}, the only unfair mechanism available for us to modify is $P(G)$. However, changing $P(G)$ will not change the actual causal effect in Eq. \ref{eq:pse_expectation}. %
\citet{zhang2016causal} bypasses this issue by only modifying the policy mechanism. However, this work requires the knowledge of an optimal unfair policy so that a fair policy can be optimized to be similar to the given unfair policy. \scales{} adopts a straightforward single-step constrained optimization approach which does not require any knowledge of the optimal, but unfair policy. Valid policies are guaranteed to satisfy the fairness principle under the world data distribution. 

An alternative approach~\cite{zhang2018fairness}  uses a family of counterfactual measures to detect different kinds of discrimination from data, and applies an explanation formula to quantitatively analyze the trade-off between outcome and procedural fairness. In practice, the explanation formula can suggest reparatory policies according to the detected discrimination. However, this framework focuses on causality-based fairness and the explanation formula cannot be used for analyzing the trade-off between outcome and non-causal fairness considerations. On the other hand, \scales{} is able to incorporate a variety of fairness principles into its problem formulation, from which optimal fair policies can be derived via planning or reinforcement learning. 

More generally, recent works in the field of ethical decision-making such as  ~\cite{svegliato2021ethically} have introduced frameworks for building ethically-compliant autonomous agents/systems. The optimization setup in  \cite{svegliato2021ethically} is similar to \scales{} in that the goal is to maximize a value function subject to moral principle (a constraint). The framework is very general in that it can represent alternative ethical theories (e.g., virtue ethics, or divine command theory), but it does not explicitly consider fairness. Here, we focus on fairness constraints and the specific constraints formulations necessary to obtain fair policies.  
The specific constraints we introduce in \scales{} and the SCMDP can potentially be incorporated into general ethical frameworks. 
\section{Case Studies} \label{sec:casestudiestitle}
In this section, we look at three case studies that showcase the various insights that \scales{} can provide when analyzing fairness principles. In particular, we will see that the choice of fairness principles and thresholds can lead to counter-intuitive policies and \scales{} can help us understand the trade-offs under alternative settings. Our case studies are conducted on the synthetic subsidized health care scenario and real-world data from  COMPAS~\cite{Mattu}, a risk assessment tool used by states across the United States of America. We utilize the CPO solver~\cite{achiam2017constrained,Ray2019} to generate the constrained policy for all our case studies. Our code is available at \url{https://github.com/clear-nus/SCALES} 

\begin{figure}
  \centering
      \includegraphics[width=0.85\linewidth]{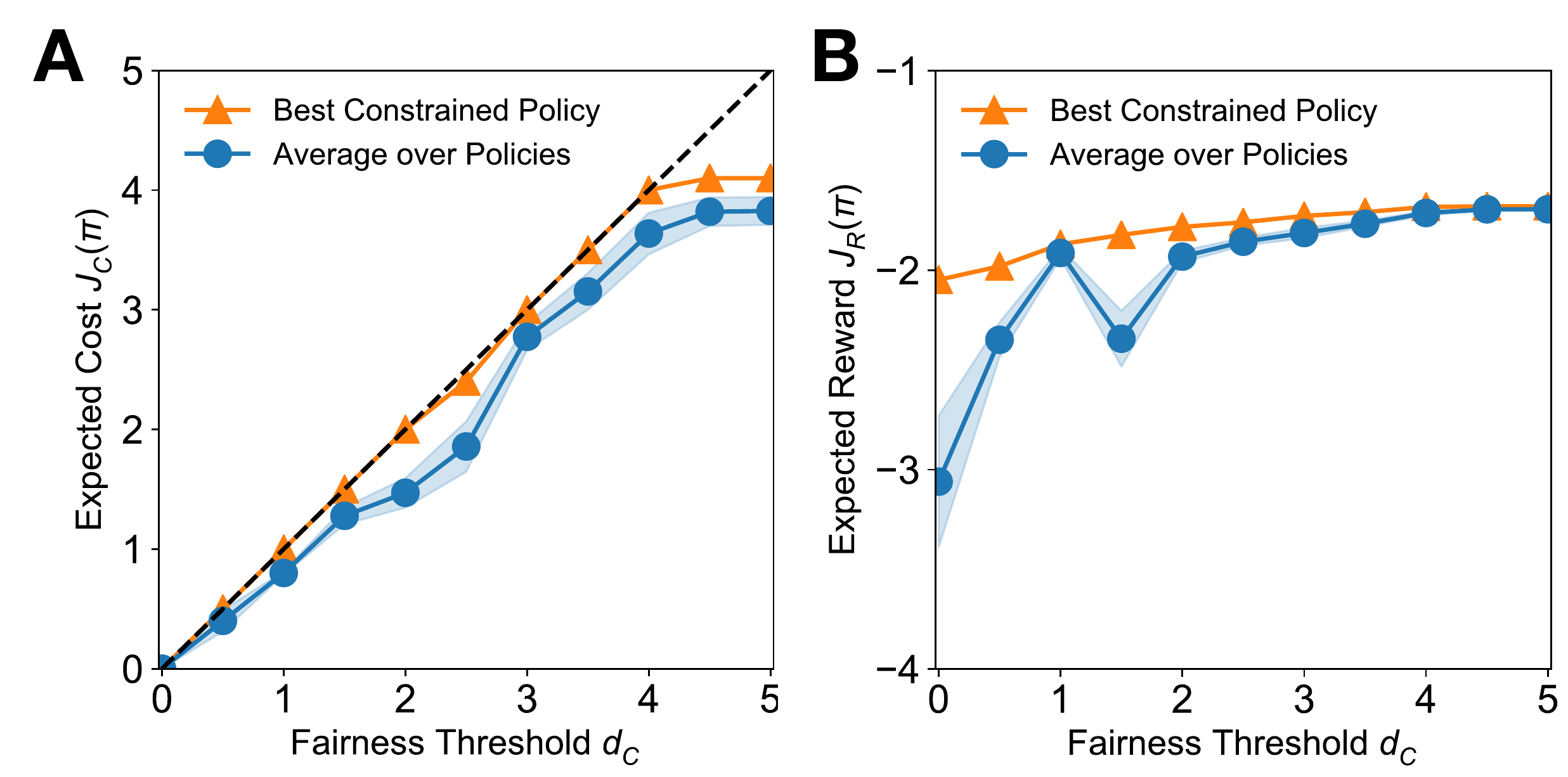}
  \caption{(\textbf{A}) Expected cost $J_C(\pi)$ of the best fair policy (highest expected reward, orange line) and averaged across the fair policies (blue line) obtained by \scales{} in Case Study 1. The shaded area indicates one standard error. These policies constrain the group-level causal effect of \textit{Gender} on \textit{Individual Benefit} (Group Outcome Fairness) to lie below the given threshold $d_C$. As the thresholds are loosened, we observe that $J_C(\pi)$ converges to $\sim4$ (similar to unconstrained policies). The best policy increases its expected cost to the specified threshold in order to maximize the expected return. (\textbf{B}) Expected return of the learned fair policies at each fairness threshold $d_C$. As the fairness threshold tightens (towards 0), the expected returns reduce which indicates that fairness is achieved at the expense of returns.}
  \label{fig:cost_reward_group}
\end{figure}

\subsection{Case Study 1: Subsidized Health Care}
\label{sec:casestudy1}

\begin{figure*}[h]
  \centering
  \includegraphics[width=1.\linewidth]{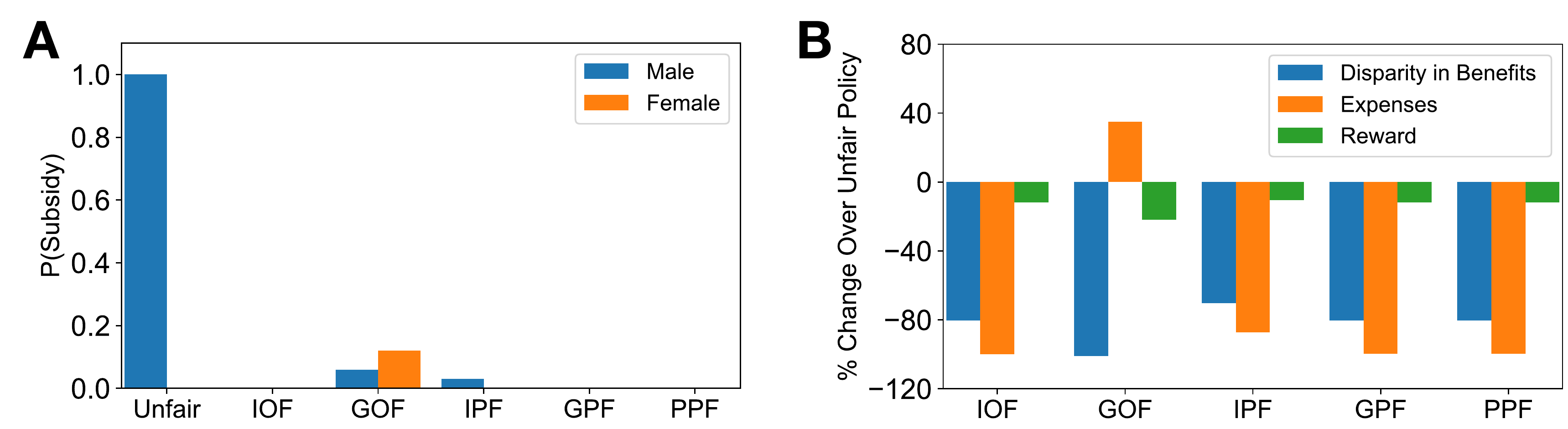}
  \caption{(\textbf{A}) Probability of providing a subsidy under various fairness principles. Under the reward-maximizing \emph{unfair} policy, a male individual is almost certainly awarded a subsidy, whilst a female individual is always denied. Imposing fairness criteria often result in undesirable policies that deny subsidies to both groups. The exception is Group Outcome Fairness (GOF). (\textbf{B}) Change in the constrained policies relative to the unfair policy. The GOF policy achieves the highest reduction in disparity, so both the male and female groups obtain similar \textit{Individual Benefits}. This comes at the cost of increased government expenses compared to the unfair policy, but the reduction in total reward appears acceptable at $\approx 20\%$.} 
  \label{fig:evaluating_policies}
\end{figure*}

For this case study, we revisit the subsidized health scenario in Figure \ref{fig:causalmodel}. In this setting, the \textit{Gender} ($G$), \textit{Health} ($H$), \textit{Employment} ($E$) and \textit{Decision} ($D$) nodes take on binary values while the remaining nodes are continuous variables. The \textit{Individual Benefit} ($I$) and \textit{Expenses} ($Ex$) are calculated using the Eqns. \ref{eq:ib_eqn} and \ref{eq:exp_eqn} respectively:
\begin{equation}\label{eq:ib_eqn}
    I =
    \begin{cases}
      3(1-E) + 5(1-H), & \text{if}\ D = 1 \text{ (Subsidy)}\\
      2(1-E) + 4(1-H), & \text{if}\ D = 0 \text{ (No Subsidy)} 
    \end{cases}
\end{equation}

\begin{equation}\label{eq:exp_eqn}
    Ex =
    \begin{cases}
      5(1-G) + 4(1-E) + 8(1-H), & \text{if}\ D = 1\\
      0  & \text{if}\ D = 0
    \end{cases}
\end{equation}
Finally, \textit{Reward} ($R$) $\triangleq (I-Ex)$. Note that the expense $Ex$ is directly affected by gender, which could result in an unfair distribution of subsidies across genders when optimizing the expected reward.  

\vspace{40px}We use \scales{} to generate fair policies under five distinct fairness principles, namely: 
\begin{enumerate}
    \item Individual Outcome Fairness (IOF)
    \item Group Outcome Fairness (GOF)
    \item Individual Procedural Fairness (IPF)
    \item Group Procedural Fairness (GPF) 
    \item Path-specific Procedural Fairness (PPF)
\end{enumerate}
These principles restrict the causal effect of the sensitive attribute $G$ on either the outcome $I$ (Outcome fairness) or decision $D$ (Procedural fairness). The causal effects are calculated either using the attributes of the current individual (Individual Fairness) or across the genders (Group Fairness). The various components of \scales{} for these fairness principles can be found in Table \ref{tab:bigtable} with the following substitutions: $\mathcal{R} = I$; $\mathcal{D} = D$; and $\mathbf{Z} = G$.

We used \scales{} to generate constrained policies under various causal fairness thresholds, $d_C$. The expected costs and rewards of these generated policies under the GOF principle are shown in Figure \ref{fig:cost_reward_group}. We observe that \scales{} successfully discovers fair policies under the given thresholds.  It is clear that tighter fairness constraints come at the cost of reduced expected rewards. This is further evident when considering the policy with the highest reward at each $d_C$ (Best Constrained Policy), which typically have an expected cost very close to the limit set by the threshold $d_C$. As the threshold values are loosened beyond 3.5, the expected cost stops increasing since the constrained policies converge to the best policy possible for this environment.

Let us turn our attention to the behavior of policies under the different fairness principles. Figure \ref{fig:evaluating_policies}(A) shows that an unfair optimal policy (that solely optimizes the expected reward) is undesirable as it only allocates subsidies to males. However, applying fairness constraints can lead to counter-intuitive behaviors---with the exception of GOF and IPF, the other fair policies simply deny subsidies to both groups in order to preserve their version of fairness. GOF, by design, attempts to reduce the \emph{Disparity in Benefits} which is defined as the difference between the $\mathbb{E}[I]$ of the male and female subpopulations. We see in \ref{fig:evaluating_policies}(A) that the GOF policy fulfills this constraint by assigning a higher probability of subsidy to the female sub-population over the males. This significantly reduces the Disparity in Benefits compared to the unfair policy. There is an increase in government expenses when following the GOF policy, but the reduction in total reward is comparable to policies obtained under other fairness principles.

\subsection{Case Study 2: Sequential Decision Making for Subsidized Health Care}

\begin{figure}
  \centering
  \includegraphics[width=0.6\linewidth]{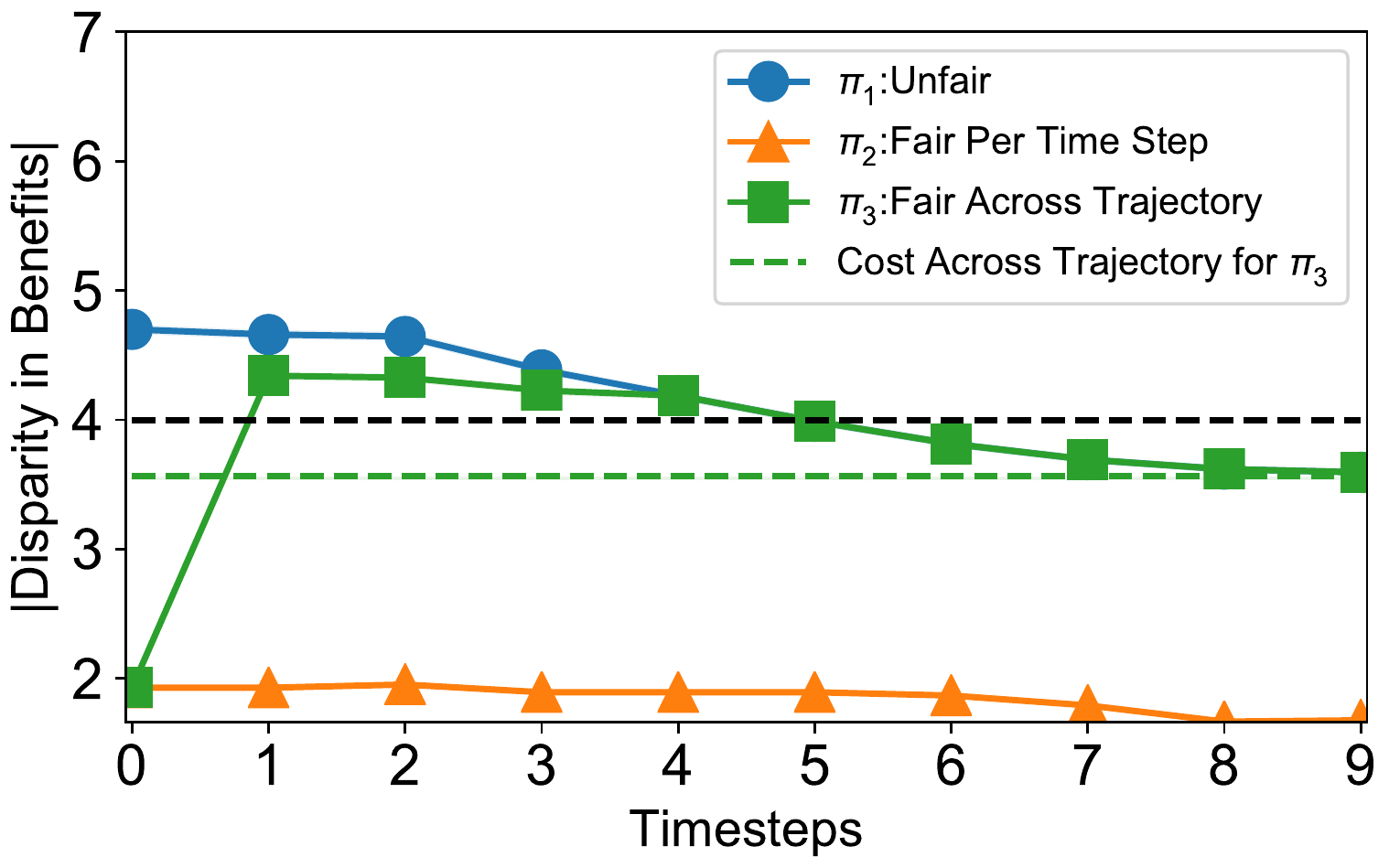}
  \caption{Effect of alternative temporal fairness constraints on the resultant \scales{} policy. $\pi_1$ is an unfair policy with \textit{disparity} (difference in individual benefits between the two stakeholders) above the threshold of 4. $\pi_2$ is a fair policy obtained by constraining the disparity at \emph{every} time step. $\pi_3$ is obtained if the fairness consideration is that the \emph{average} disparity across the trajectory is constrained. The average cost of $\pi_3$ (dashed green line) is below the threshold but the disparity may breach the threshold at any given time step. \scales{} enables the comparison of different temporal fairness considerations.}
  \label{fig:seq_setting}
\end{figure}

In this case study, we extend the health care scenario in the previous section to a sequential decision-making problem with a horizon of 10. %
We consider \textit{Health} as a continuous variable whose value changes based on the decision to subsidize. In addition, the decision-maker must now decide to subsidize the healthcare of two stakeholders at every trajectory, which prompts us to consider  Distributive Fairness. In our scenario, the total $I$, $Ex$, and $R$ at each step are obtained by summing up the corresponding values of both the stakeholders. For the rest of the section, we denote the time step by a superscript and the stakeholder via a subscript.

In sequential decision-making settings, fairness definitions must also take into consideration the temporal dimension. For instance, when trying to constrain the \emph{Disparity in Benefits} ($DiB$) between the two stakeholders, we can define $DiB$ using either of the following definitions:  
\begin{align}
&DiB_{ps} \triangleq \max_{t}[I^t_1 - I^t_2 ]\label{eq:fairps}\\
&DiB_{ac} \triangleq \frac{1}{10}\sum_{t}[I^t_1 - I^t_2 ]\label{eq:fairacr}
\end{align}
In words, constraining the absolute value of $DiB_{ps}$ ensures that the disparity remains under a given threshold at every time step. We call this fairness principle \textit{Fairness Per Time Step} (FPTS). Meanwhile, constraining the absolute value of $DiB_{ac}$ ensures that the average disparity value across the entire trajectory is below a given threshold. We call this fairness principle \textit{Fairness Across Trajectory} (FAcT). These two fairness principles could lead to different policy behaviors.

Figure \ref{fig:seq_setting} shows the behavior of the two fairness principles mentioned above. As expected, the absolute value of the disparity lies below the given threshold of +4 at every time step when following the FPTS policy. The dotted green line shows the absolute value of $DiB_{ac}$ as per Def. \ref{eq:fairacr}. While the average disparity remains under the threshold of 4 when following the fair policy under FAcT, the same cannot be said for the disparity at each time step --- as shown in Figure \ref{fig:seq_setting}, the disparity value exceeds the given threshold at some time steps under the FAcT policy.

 \begin{figure}
  \centering
  \includegraphics[width=0.5\linewidth]{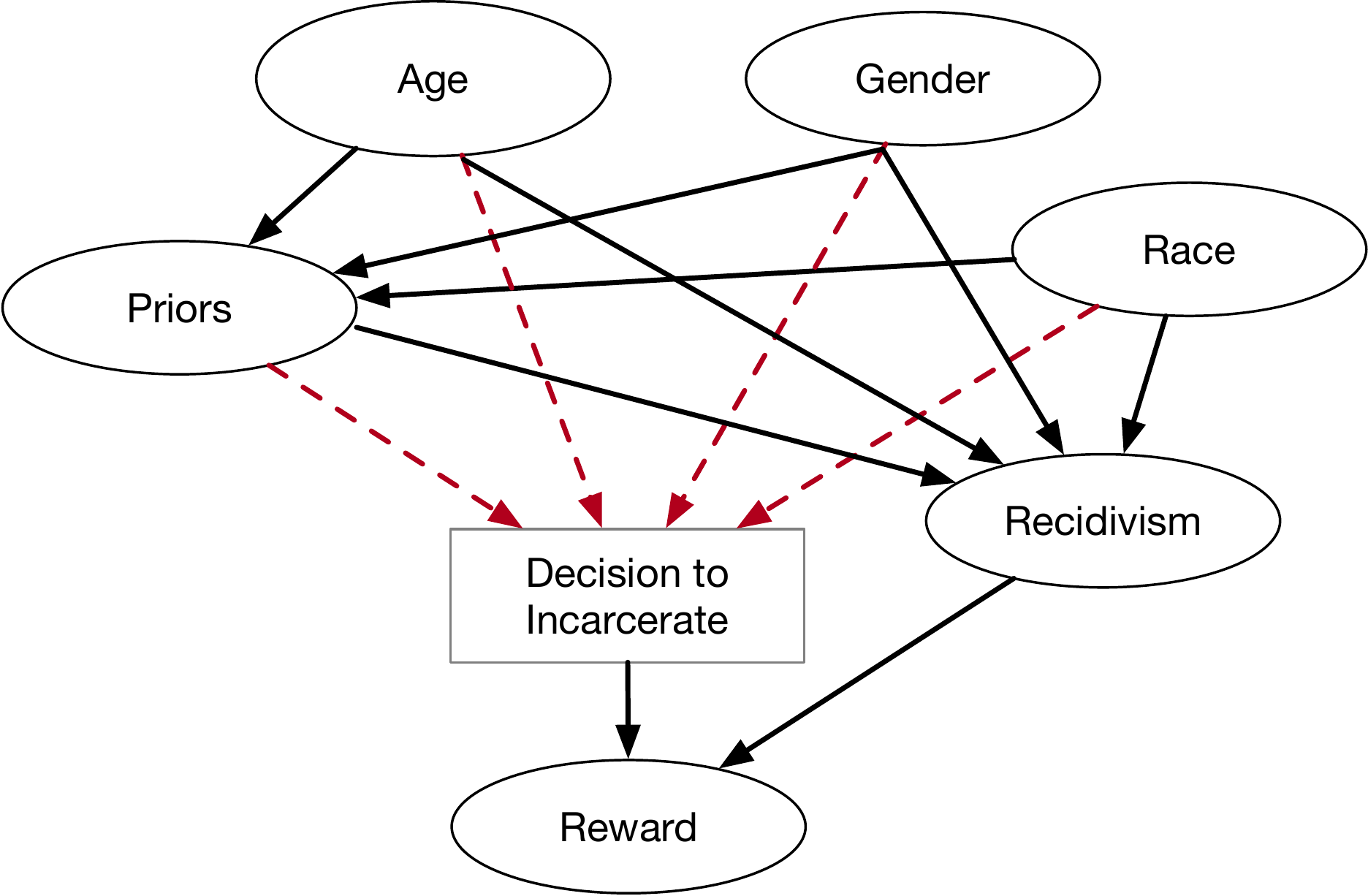}
  \caption{Causal Model for the COMPAS Case Study.}
  \label{fig:compas_scm}
\end{figure}

 \begin{figure}
  \centering
  \includegraphics[width=1.\linewidth]{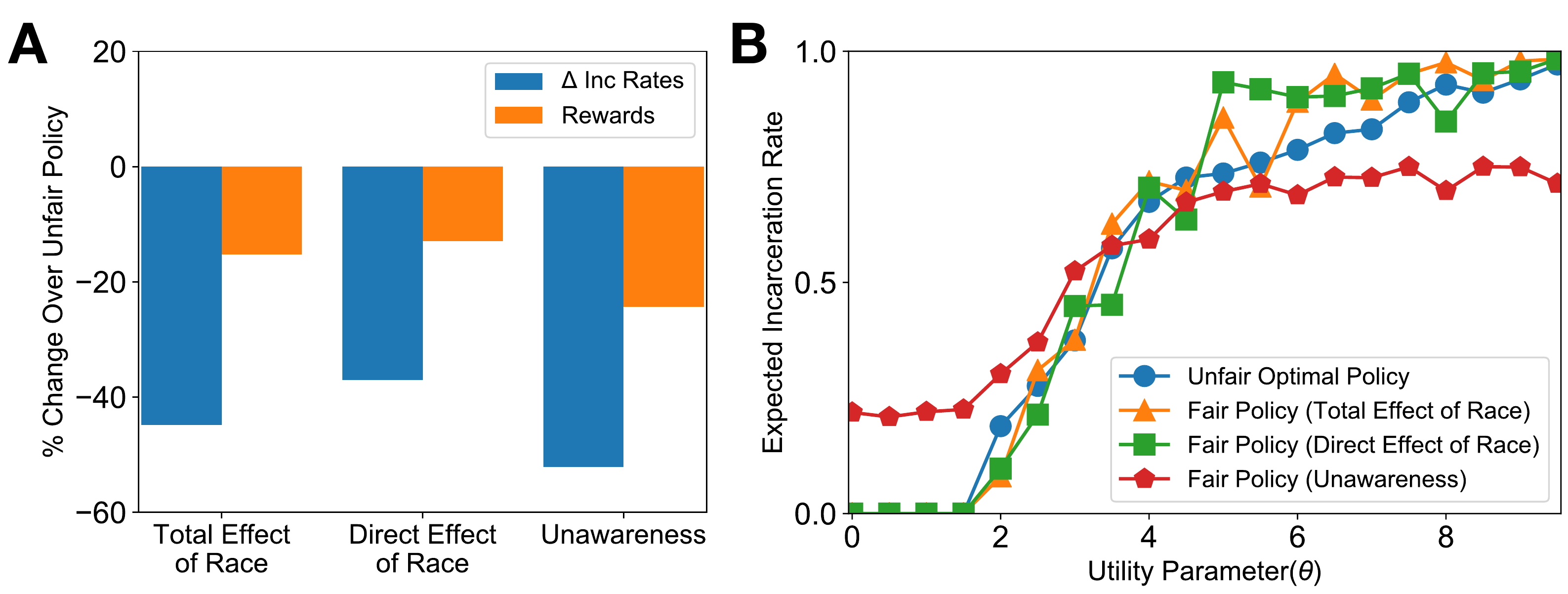}
  \caption{(\textbf{A}) Expected return for the unfair policy and policies learned under various fairness principles. Higher utility parameter $\theta$ implies a high penalty when an individual recidivates while out on bail. Regardless of fairness principles, the incarceration rates increase as the utility parameter is increased. (\textbf{B}) Relative change (negative value indicates a reduction) of the difference in incarceration rates between \textit{Race 0} and \textit{Race 1} when compared to the unfair policy when evaluated at $\theta=5.0$. When fairness is imposed, the difference in incarceration rates decreases. Among the compared fairness criteria, fairness through unawareness results in the largest reduction in incarceration rates, but it also suffers the highest loss of reward.}
  \label{fig:compas_rates}
\end{figure}

\subsection{Case Study 3: COMPAS Risk Assessment}

In this final case study, we evaluated \scales{} on data generated by the COMPAS~\cite{Mattu} risk assessment tool. Since COMPAS is widely used by the criminal justice system, the decisions generated by this tool have real-world consequences. 

COMPAS assesses an individual, identified by \textit{Race, Gender, Age} and \textit{Prior Charges}, and decides if the individual should be detained before their trial, generating a \textit{Score}. We follow the pre-processing steps in \cite{nabi2019learning} (with associated SCM in Fig. \ref{fig:compas_scm}) and binarize these variables. A \textit{Score} of 1 indicates the person is detained while 0 indicates the person is released on bail. The reward $R$ obtained depends on the occurrence of recidivism $V$ in the next two years,
\begin{equation}
    R=
    \begin{cases}
      -1, & \text{if}\ Score = 1\\
      +1, & \text{if}\ Score = 0 \text{ and } V = 0\\
      -1*\theta, & \text{if}\ Score = 0 \text{ and } V = 1
    \end{cases}
\end{equation}
where $V=1$ indicates that recidivism has occurred and $V=0$ indicates otherwise. $\theta$ is a utility parameter that adjusts the risk of granting bail to a given individual. 

We compare the effect of various principles that attempts to limit the causal effect of the sensitive attribute \textit{Race} on the COMPAS \textit{Score}. In particular, we investigate three different fairness considerations:
\begin{enumerate}
    \item Limiting the total causal effect of \textit{Race} on \textit{Score};
    \item Limiting the causal effect of \textit{Race} on \textit{Score} through only the direct path;
    \item Ignoring \textit{Race} as an input to the policy (Fairness through unawareness);
\end{enumerate}

Figures \ref{fig:compas_rates}(A) and \ref{fig:compas_rates}(B) summarize our results. We observe that the expected incarceration rate of various policies increases with the utility parameter $\theta$. This occurs because as $\theta$ increases, so does the risk associated with allowing a person to go on bail. Figure \ref{fig:compas_rates}(B) compares the percentage change of the disparity in Incarceration Rates between \textit{Race}=0 and \textit{Race}=1. All fairness principles achieve a smaller disparity compared to the unfair policy. Even though the \emph{Fairness through Unawareness} policy achieves the largest reduction in disparity, it also incurs the largest loss in reward when compared to the unfair policy. %

\section{Conclusion}

In this paper, we present \scales, a decision-theoretic framework that is able to represent different kinds of fairness principles. We discuss how to translate common fairness principles into a standard form, which results in a specific form of CMDPs. We illustrate how \scales{} can provide both fair policies and corresponding insights through several case studies on a synthetic health subsidy scenario, as well as on the real-world COMPAS dataset. 

Looking forward, the cost functions associated with the various fairness principles evaluated in our work can be policy-dependent (since we need to reason how the current policy affects individuals or groups in the given task). While the standard CPO formulation assumes policy-independent cost functions, the solver remains performant in our case studies. That said, performance may vary across domains and developing a dedicated solver for policy-dependent fairness constraints may be interesting future work. In addition, we are currently looking to apply \scales{} in more complex domains such as human-robot interaction~\cite{houston2022,brandao2020fair}. %

\section*{Acknowledgments}
This research is supported by the National Research Foundation Singapore under its AI Singapore Programme (Award Number: AISG2-RP-2020-017).

\bibliographystyle{plainnat}
\bibliography{references}

\end{document}